# An unsupervised cluster-level based method for learning node representations of heterogeneous graphs in scientific papers

Jie Song, Meiyu Liang*, Zhe Xue, Junping Du, Kou Feifei

(*Beijing Key Laboratory of Intelligent Communication Software and Multimedia, School of Computer, Beijing University of Posts and Telecommunications (National Demonstration Software School), Beijing 100876*)

(songs@bupt.edu.cn)

**Abstract** Learning knowledge representation of scientific paper data is a problem to be solved, and how to learn the representation of paper nodes in scientific paper heterogeneous network is the core to solve this problem. This paper proposes an unsupervised cluster-level scientific paper heterogeneous graph node representation learning method (UCHL), aiming at obtaining the representation of nodes (authors, institutions, papers, etc.) in the heterogeneous graph of scientific papers. Based on the heterogeneous graph representation, this paper performs link prediction on the entire heterogeneous graph and obtains the relationship between the edges of the nodes, that is, the relationship between papers and papers. Experiments results show that the proposed method achieves excellent performance on multiple evaluation metrics on real scientific paper datasets.

CLC number TP391

There are various types of edges (relationships) between heterogeneous graph nodes[1], and the different attributes of each edge will also lead to the distance between nodes[2]. The current difficulty in processing heterogeneous graphs is that on the one hand, the structural information of the graph must be processed[3], and on the other hand, the attributes of each node must be paid attention to[4]. Traditional machine learning methods[5] focus on the features of individual nodes while ignoring structural information[6]. Graph neural networks learn new feature vectors for nodes through a recursive neighborhood aggregation strategy, and many supervised graph neural network models have been proposed for graph data representation learning. However, labeled data in heterogeneous graphs of scientific papers are not always available, and these algorithms are not suitable for unsupervised learning of heterogeneous graphs of scientific papers.

In heterogeneous graph research, meta-paths[7] have been widely used to represent composite relationships with different semantics. UCHL uses the structure of meta-paths to model the connection semantics in heterogeneous graphs. Based on different meta-paths, heterogeneous graphs Decomposed into isomorphic graphs with specific semantics, applying graph convolutional neural networks to capture local representations of nodes with specific semantics, and aggregating node representations with different semantics based on an attention mechanism, by maximizing local-global, local-cluster center mutual information and learn representations embedded with graph-level semantic structure information without relying on any supervised label guidance information. The main contributions of this paper are as follows:

1) An unsupervised cluster-level based method for learning the node representation of the heterogeneous graph of scientific papers, UCHL, is proposed, which acquires the knowledge representation of scientific paper data by learning the node representation of the paper data in the heterogeneous network of scientific papers;

2) Learning the low-dimensional embedding space

This work is supported by National Key R&D Program of China (2018YFB1402600), the National Natural Science Foundation of China (61772083, 61877006, 61802028, 62002027).
**Corresponding author**: Meiyu Liang (meiyu1210@bupt.edu.cn)

representation of heterogeneous graph nodes in scientific papers based on mutual information theory and cluster center, while retaining the semantic structure information and node feature information of the graph;

## 1 Related work

Deepwalk[8] belongs to one of the graph embedding techniques that use walks to traverse the graph by moving from one node to another. Node2vec[9] was one of the first deep learning attempts to learn from graph structured data, considering the Breadth-First Search (BFS) and Depth-First Search (DFS) processes. graph2vec[10] essentially learns to embed a set of subgraphs of a graph with a user-specified number of edges, which is passed to a neural network[11] for classification. SDNE[12] attempts to learn from two different metrics: first-order proximity and second-order proximity. LINE[13] explicitly defines two functions; one for first-order approximation and the other for second-order approximation. HARP[14][15] improves the solution and avoids local optima through better weight initialization, and uses graph coarsening techniques to aggregate related nodes into "super nodes", essentially a graph preprocessing step that simplifies graph to speed up training.

GCN[16] is a multi-layer graph convolutional neural network that encodes adjacency $A$ and feature matrices $X$ as embeddings $H$. GATs[17] uses attention coefficients instead of Laplacian matrices on the basis of GCN, which better integrates the correlation between vertex features into the model. Various GCN variants have been proposed to address the problem of learning graph embeddings. GraphSAGE[18] samples the neighbor vertices of each vertex in the graph. GCN-LPA[19] analyzes the theoretical relationship between Graph Convolutional Networks (GCN) and Label Propagation Algorithms (LPA). CNMPGNN[20][21] is a topic based on common neighbors that generalizes and enriches structural patterns. ACM-GCN[22] considers the influence of graph structure and input features on GNNs, and proposes an Adaptive Channel Mixing (ACM) framework to adaptively exploit aggregation, diversity and identity channels in each GNN layer to address harmful heterogeneity Addiction.

Metapath2vec[23] is a meta-path-based heterogeneous graph embedding method, which can only deal with a specific meta-path. ESim[24][25] can exploit multiple meta-paths, but fails to understand the importance of meta-paths. HAN[26] learns the importance of neighbors and multiple meta-paths based on an attention mechanism. MAGNN[27] considers intermediate nodes in meta-paths, aggregating intra-meta-path and inter-meta-path information. HetGNN[28][29] adopts a random walk strategy to sample neighbors, using a specialized Bi-LSTM to integrate heterogeneous node features and neighboring nodes. HetSANN[30] learns different types of adjacent nodes as well as related edges through a type-aware attention layer. Based on Transformer[31] The architecture of HGT[32] Learn the characteristics of different nodes and their relationship to specific types of parameters. DGI[33] describes how the DIM[34][35] idea of maximizing mutual information is applied to the graph domain. On the basis of DGI, HDGI[36] is the first unsupervised method to apply maximizing mutual information in heterogeneous graph representation learning. GIC[37] is an unsupervised graph representation learning method using cluster-level[38] node information, and it is also an extension of DGI work, but it is applied to homogeneous graphs.

## 2 An unsupervised cluster-level based method for learning node representations of heterogeneous graphs in scientific papers

In order to learn the knowledge representation of scientific paper data, this paper proposes an unsupervised cluster-level based learning method of node representation in heterogeneous graph of scientific papers (UCHL) to obtain the representation of nodes in heterogeneous graph of scientific papers. The overall framework of the method is as follows: Figure 1.

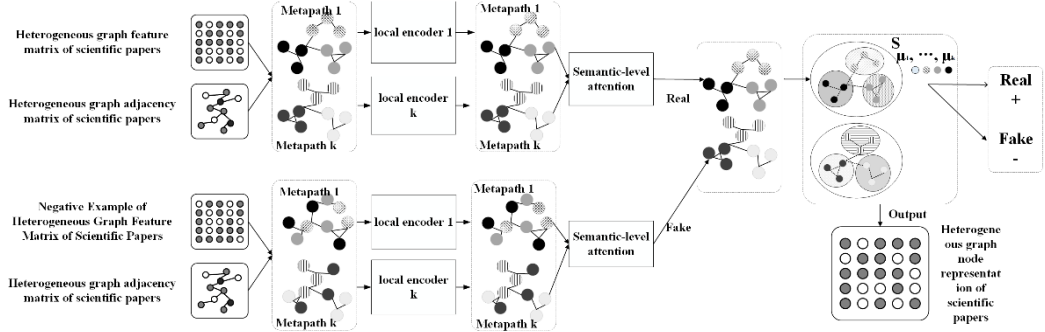

Fig. 1Framework of Unsupervised Cluster-level Scientific Paper Heterogeneous Graph Node Representation Learning

## 2.1 Problem Definition

The heterogeneous graph of scientific papers is denoted as $G = (V, E)$, which consists of an object set $V$ and a connection set $E$. Heterogeneous graphs are also associated with node type mapping functions $\varphi: V \to A$ and connection type mapping functions $\omega: E \to B$. $A$ and $B$ represents a collection of predefined object types and connection types, where $|A| + |B| > 2$. Based on heterogeneous graphs $G$ and node feature sets $X$ of scientific papers, the goal of UCHL is to learn a $d$ dimensional node representation $H \in R^{|V| \times d}$ that $H$ includes graph structure information and node feature information. Here we only focus on $V$ the representation learning of scientific paper nodes.

In the heterogeneous graph of scientific papers, two adjacent nodes may be connected by different types of edges. These edges are defined as meta-paths, which represent the node types and the number of edges between two adjacent nodes in the heterogeneous graph of scientific papers. Types of. Therefore, the set of meta-paths is defined as $\{\phi_1, \phi_2, ..., \phi_k\}$, which $\phi_i$ represents the $i$ type of meta-path. For example, in the heterogeneous graph of scientific papers, the edge relationship between two papers of the same author and the edge relationship between two papers of the same discipline is Different meta paths. For meta-paths $\phi_i$, if $v_m \in V$ a meta-path relationship exists between nodes $v_m$ and then, $v_n \in V$ sum $v_n$ is based $\phi_i$ on adjacent nodes. Such information can be represented by a meta-path adjacency matrix $A^i \in R^{|V| \times |V|}$, if $A^i_{mn} = A^i_{nm} = 1$ it means that $m$ the $n$th node is connected to the th node, otherwise it means that it is not connected.

## 2.2 UCHL overall architecture

The overall structure of UCHL is shown in Figure 1. Representation learning is performed for all specific meta-path nodes, the adjacency matrix under each meta-path is calculated $A^i$, and the corresponding negative sample data is constructed, and they are simultaneously input into the local encoder for encoding. The meta-path local encoder is a hierarchical structure, Individual node representations are learned separately from each meta-path-based adjacency matrix, and then aggregated through semantic-level attention to obtain the output node representation of the meta-path-based local representation encoder $H$, which in turn obtains a global graph representation $s \in R^{1 \times d}$ as well as clusters $\mu_c$.

## 2.3 Construction of Negative Samples for Heterogeneous Graphs of Scientific Papers

Negative samples are used to construct nodes that do not exist in the original graph. In the heterogeneous graph $G$ of scientific papers, rich and complex structural information is obtained from the meta-path-based adjacency matrix set. Eq. (1) illustrates the process of negative sample generation:

$$(\tilde{X}, \{A^1, ..., A^k\}) = C(X, \{A^1, ..., A^k\}) \quad (1)$$

where $C$ represents the random shuffling function, keeping all meta-path-based adjacency matrices unchanged, changing the indices of nodes to break node-level connections between them. The structure of the whole graph has not changed, but the initial features corresponding to each node have changed.

## 2.4 Hierarchical Local Encoders for Heterogeneous Graphs of Scientific Papers

The meta-path-based hierarchical local encoder has a two-layer structure. This paper first derives a node representation from each meta-path-based adjacency matrix $A^i$ ($i$=1, 2, ..., $k$) separately, and then aggregates the node representations under all meta-paths through an attention mechanism. A graph attention neural network is used as a hierarchical local encoder to derive a node representation containing initial node features $X$ and adjacency matrices $A^i$ ($i=1, 2, \ldots, k$), and the $H^i$ learned hidden representation of nodes is shown in Eq. (2):

$$\vec{h}^i_p = \mathop{\|}_{k=1}^{K} \sigma(\sum_{q \in N^i_p} \alpha^i_{pq} W \vec{x}_j) \qquad (2)$$

where $W$ is the shared specific meta-path linear weight matrix transformation, $N^i_p$ is the $p$ the set of neighbor nodes that the node is $\alpha^i_{pq}$ based on, is $\phi_i$ the attention weight $K$ based on $\phi_i$ the $q$ two connected nodes, and $p$ is the number of heads in the multi-head attention mechanism.

After the learning of the hierarchical local encoder, the meta-path node representation set based on different semantics can be obtained $\{H^1, H^2, \ldots, H^k\}$. The representations learned by nodes based on each meta-path only contain semantically specific information in the heterogeneous graph. To aggregate the representations of nodes, a semantic attention layer is added to learn the weights that should be assigned to each meta-path. The $\{S^1, S^2, \ldots, S^k\}$ semantic attention layer learns the attention weight is guided by the binary cross-entropy loss of whether the node belongs to the original graph.

To make representations based on different meta-paths comparable, a linear transformation needs to be used to transform the representation of each node, defined by a shared weight matrix $W_{sem}$ and shared bias vector $b$ parameter. The importance of representations based on different meta-paths will be $q$ measured by sharing the attention vector. The importance of the meta-path $\phi_i$ can be calculated by Eq. (3):

$$e^i = \frac{1}{N} \sum_{j=1}^{N} \tanh(q^T \cdot [W_{sem} \cdot h^i_j + b]) \qquad (3)$$

According to the importance of the meta-path, this paper will use the $softmax$ function to normalize it, as shown in Eq. (4):

$$S^i = soft\max(e^i) = \frac{\exp(e^i)}{\sum_{j=1}^{k} \exp(e^j)} \qquad (4)$$

The weights of different meta-paths are used as coefficients for linear combination to calculate the final graph global representation H, corresponding to Eq. (5):

$$H = \sum_{i=1}^{k} S^i \cdot H^i \qquad (5)$$

and the fake graph node representation $\tilde{H}$ through the hierarchical local encoder and semantic level attention $H$, the global graph representation $s \in R^{1 \times d}$ is calculated by Eq. (6) for all node representations.

$$s = \sigma(\frac{1}{N} \sum_{i=1}^{N} h_i) \qquad (6)$$

where $\sigma$ represents the logical sigmoid function, $N$ is the number of nodes, and $h_i$ represents the representation of each node. Cluster clusters are $\mu_r$ first computed by clustering the fine-grained representation and then computing the average of all nodes within the cluster. $\mu_r$ ($r = 1, 2, \ldots,$) are $R$ obtained by layers implementing a differentiable version of K - means clustering. In the final optimization, $\mu_r$ the update of can be iteratively carried out by Eq. (7) and Eq. (8):

$$\mu_r = \frac{\sum_i c_{ir} h_i}{\sum_i c_{ir}} \qquad (7)$$

$$c_{ir} = \frac{\exp(-\beta sim(h_i, \mu_k))}{\sum_j \exp(-\beta sim(h_i, \mu_j))}, j = 1, \ldots, R \qquad (8)$$

where $sim(*,*)$ represents the similarity function between two instances and $\beta$ is a hyperparameter that tends to infinity giving each cluster cluster assignment a binary value.

To estimate and maximize mutual information, it is $i$ computed according to the cluster cluster to which each node belongs $z_i$, which represents the corresponding cluster representation of each node. The mutual information between the sums $z_i$ of each node can then be maximized. $h_i$ In order to calculate each node $i$'s $z_i$, this paper applies $i$ the weighted average of the representation of the cluster to which the node belongs, as shown in Eq. (9):

$$z_i = \sigma(\sum_{r=1}^{R} c_{ir} \mu_r) \qquad (9)$$

Among them $c_{ir}$, it has the same meaning as in Eq. (8), it represents the degree to which the node is assigned to the cluster $r$, it is a soft assignment value, and it $\sigma$ represents the logical sigmoid function.

### 2.5 Mutual Information Agent for Heterogeneous Graphs of Scientific Papers

A discriminator is defined as a proxy for estimating mutual information by assigning higher scores to positive examples than negative examples, positive examples are obtained by pairing node representations $h_i$ with node cluster representations in the real graph, and node representations are obtained by pairing node representations $z_i$ with node cluster representations in the fake graph. $\tilde{h}_i$ Clusters represent $z_i$ pairings to obtain negative examples. As a discriminator function $D$, it is a proxy for estimating the mutual information between the node representation and the graph global representation, using a bilinear scoring function as shown in Eq. (10):

$$D(h_i, z_i) = \sigma(h_i^T z_i) \qquad (10)$$

where $\sigma$ represents the logistic sigmoid nonlinear function.

The discriminator is a standard binary cross-entropy loss function whose purpose is to maximize the product of the expected log ratio of the samples in the joint distribution (positive examples) and the marginal distribution (negative examples). The positive example is the $s$ AND pairing $h_i$ of the real input graph $G$, but the negative example is the $s$ AND $\tilde{h}_i$ pairing of the graph, and considering the information of the clusters, $D$ the loss function of the discriminator consists of two parts, one of which $L_g$ is between the global representation of the graph and the node representation. The cross-entropy loss $L_c$ is the cross-entropy loss between the cluster representation and the node representation, as shown in Eq. (11) and Eq. (12), respectively:

$$L_g = \frac{1}{2N}(\sum_{i=1}^{N} E_{(X,A)}[\log D(h_i, s)] \\ + \sum_{i=1}^{N} E_{(\tilde{X},\tilde{A})}[\log(1 - D(\tilde{h}_i, s))]) \qquad (11)$$

$$L_c = \frac{1}{2N}(\sum_{i=1}^{N} E_{(X,A)}[\log D(h_i, z_i)] \\ + \sum_{i=1}^{N} E_{(\tilde{X},\tilde{A})}[\log(1 - D(\tilde{h}_i, z_i))]) \qquad (12)$$

The total loss function of UCHL is shown in Eq. (13).

$$L = \theta L_g + (1-\theta)L_c \qquad (13)$$

where $\theta \in [0,1]$ is a hyperparameter. In this paper, by continuously optimizing the loss in Eq. (13), an optimal solution can be obtained, and the node representation of the heterogeneous graph of scientific papers can be learned. The algorithm flow is shown in Table 1.

**Table 1 Flowchart of UCHL algorithm**

**Input:**
　　Heterogeneous graph feature matrix of scientific papers $X$
　　Heterogeneous graph adjacency matrix of scientific papers $A$
**Output:**
　　Heterogeneous graph node representation of scientific papers $H$
1.　Construct the corresponding negative sample data
2.　Hierarchical local encoders learn node representations under each meta-path
3.　Aggregate node representations under all meta-paths through an attention mechanism
4.　Semantic attention layers to learn the weights that each meta-path should assign $S^1, S^2, \dots, S^k$
5.　Learning Final Node Representations via Mutual Information Agents of Heterogeneous Graphs

## 3 Experimental results and analysis

### 3.1 Dataset

In order to verify the proposed unsupervised cluster-level based learning method of heterogeneous graph node representation (UCHL), a comparative test with other related algorithms is carried out on three real public datasets.

DBLP dataset: contains 14328 papers (P), 4057 authors (A), 20 conferences (C), 8789 terms (T). The authors are divided into four areas: databases, data mining, machine learning, and information retrieval. This paper uses the meta-path set {APA, APCPA, APTPA} to conduct experiments.

ACM dataset: Extract papers published in KDD, SIGMOD, SIGCOMM, MobiCOMM and VLDB, and classify papers into three categories (database, wireless communication, data mining). This paper constructs a heterogeneous graph including 3025 papers (P), 5835 authors (A), and 56 topics (S). Experiment with the meta-path set {PAP, PSP}.

AMINER dataset: contains 4472 authors (A), 7,623 papers (P), 101 conferences (C) and 10 paper classifications (L). This paper uses the meta-path set {APA, APCPA, APTPA} to conduct experiments.

**Table 2 Heterogeneous Graph Dataset Details**

| data set | AB relationship | A quantity | B quantity | Number of AB relationships | meta path |
|---|---|---|---|---|---|
| DBLP | PA | 14328 | 4057 | 19645 | APA |
| | P -C | 1 4328 | 20 | 14328 | APCPA |
| | P -T | 14327 | 8789 | 88420 | APTPA |
| ACM | P-A | 3025 | 5835 | 9744 | PAP |
| | P-S | 3025 | 56 | 3025 | PSP |
| AMI NER | P-A | 7623 | 4472 | 15213 | APA |
| | P-C | 7623 | 101 | 4158 | APCPA |
| | P - L | 7 623 | 1 0 | 7 623 | APLPA |

3.2 Comparison method

This paper compares the proposed UCHL with the following unsupervised representation learning methods:

***Raw Feature***: Directly use node text features;

***DeepWalk***: a random walk-based network designed for isomorphic graphs;

***Metapath2vec***: Based on meta-paths, but can only handle a specific meta-path;

***ESim***: Capture semantic information from multiple meta-paths.

At the same time, this paper also compares the following supervised representation learning methods:

***GCN***: Semi-supervised method for node classification in isomorphic graphs;

***GAT***: A supervised representation learning method based on attention mechanism;

***HAN***: Employ node-level attention and semantic-level attention to capture information from all meta-paths.

In the experiments, some edges are hidden in the input graph, and the goal is to predict the existence of these edges based on the computed node representations, and $i$ the probabilities of edges between $\sigma(h_i^T h_j)$ nodes are $j$ given by, where $\sigma$ is a logistic sigmoid function. Set the edges of 5% positive samples and negative samples as the validation set, and set the edges of 10 % positive samples and negative samples as the test set, and the learned representation feature dimension $d$ =16. We tested the area under the ROC curve, the AUC score, which is equal to the probability that a randomly selected edge ranks higher than a randomly selected negative edge and the average precision AP score, which is the area under the precision - recall curve; here, the precision is given by TP /(TP+FP) is given and recall is given by TP/(TP+FN). Among them, TP represents the number of positive cases predicted as positive cases, FP represents the number of negative cases predicted as positive cases, and FN represents the number of positive cases predicted as negative cases.

3.3 Experimental results and analysis of UCHL

This section presents the performance comparison and parameter analysis of the proposed UCHL and other methods.

**Table 3 Performance Comparison of UCHL a and Other Comparative Methods on DBLP**

| | AUC | AP |
|---|---|---|
| Raw Feature | 70.24 | 69.73 |
| DeepWalk | 7 5.53 | 7 4.69 |
| Metapath2vec | 7 9.65 | 7 9.12 |
| ESim | 8 7.45 | 8 6.91 |
| GCN | 77.56 | 76.14 |
| GAT | 84.34 | 83.28 |
| HAN | 90.67 | 8 9.53 |
| UCHL | **93.26** | **93.11** |

The experimental results of the comparison on the DBLP dataset are shown in Table 3. Since the proposed UCHL method fully considers the impact of different meta-paths on the overall results, the proposed UCHL achieves better performance than all comparison methods in terms of AUC and AP. Compared with the second-ranked method HAN, UCHL achieves 2.59% and 3.58% improvement in AUC and AP, respectively. Compared with using the original features directly, UCHL has more than 20 % performance improvement in both AUC and AP metrics.

**Table 4 Performance Comparison of UCHL and Other Comparative Methods on ACM**

| | AUC | AP |
|---|---|---|
| Raw Feature | 69.98 | 68.22 |
| DeepWalk | 7 2.82 | 7 1.97 |
| Metapath 2 vec | 7 6.27 | 7 3.71 |
| eSim | 8 8.37 | 8 6.13 |
| GCN | 78.02 | 7 7.36 |
| GAT | 84.68 | 82.87 |
| HAN | 8 7.26 | 8 7.11 |
| UCHL | **92.48** | **91.69** |

Table 4 shows the experimental results of UCHL and other comparison methods on the ACM dataset. It can be observed that UCHL still achieves the best performance, which is about 20 % higher than the original features or traditional methods such as DeepWalk. Compared with the best supervised method HAN and unsupervised method ESim, the AUC and AP of UCHL are improved by 5.22%, 4.11% and 4.58%, 5.56% respectively.

**Table 5 Performance Comparison of UCHL and Other Comparative Methods on AMINER**

|  | AUC | AP |
|---|---|---|
| Raw Feature | 66.16 | 63.91 |
| DeepWalk | 68.94 | 67.39 |
| Metapath 2 vec | 71.39 | 68.97 |
| eSim | 83.68 | 81.61 |
| GCN | 74.33 | 73.31 |
| GAT | 79.62 | 79.17 |
| HAN | 83.14 | 82.90 |
| UCHL | **85.43** | **83.97** |

The experimental results on the AMINER dataset are shown in Table 5. As can be seen from Table 5, the performance of the proposed UCHL on AUC and AP once again surpasses all comparison methods, in which, based on the original features, UCHL achieves nearly 17% on AUC and nearly 20% on AP, respectively. Compared with the suboptimal method HAN, UCHL improves AUC by 2.29 % and AP by 1.07%. A hyperparameter is provided $\theta$ to tune the effect of the two-part loss on the overall performance of the model. In order to verify the influence of hyperparameters in Eq. (14) $\theta$ on UCHL, parameter experiments were carried out on three datasets respectively, and the experimental results are shown in Figures 2 and 3. Search $\theta$ in [0.1, 0.9], step size is set to 0.1, perform 9 experiments with different hyperparameters $\theta$, record the results of the experiments on three datasets, and observe how the AUC and AP of UCHL will change.

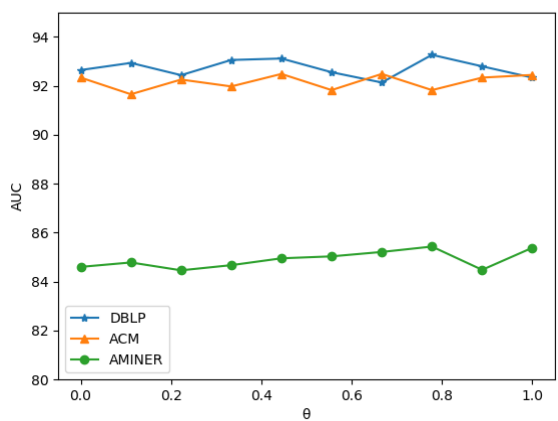

Fig. 2 AUC values of UCHL for different $\theta$ on three datasets

It can be observed from Figure 2 that the AUC performance of UCHL on the three datasets DBLP, ACM and AMINER remains almost unchanged as it varies from 0.1 to 0.9, that is to say, the change of hyperparameters has no effect on the proposed method has little effect in the end.

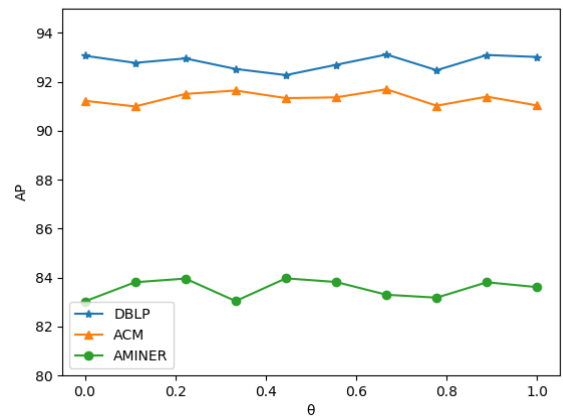

Fig. 3 AP values of UCHL for different $\theta$ on three datasets

Figure 3 shows that the AP performance of UCHL on the DBLP, ACM and AMINER datasets $\theta$ remains stable with fluctuations, and the optimal performance is generally obtained in $\theta$ the [0.4-0.7] interval. Overall, UCHL is not sensitive to parameters. The effect of the number of clusters $R$ on UCHL was verified. Taking the ACM dataset as an example, the learned technical papers are represented by TSNE dimensionality reduction and the 2D spatial graph node representation is drawn. Since the papers in the ACM dataset are divided into three categories, we set the number of categories when drawing the TSNE dimensionality reduction graph as 3, we take 3, $R$ 4, and 5, respectively, and draw the results shown in Figure 4.

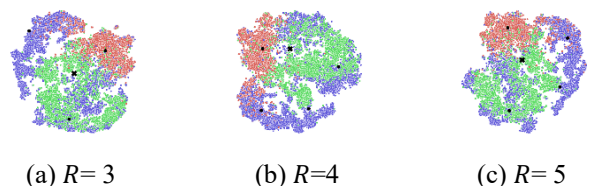

(a) $R$= 3          (b) $R$= 4          (c) $R$= 5

Fig. 4 Effects of different cluster numbers $R$ on the quality of node representation learned by UCHL

The results were evaluated using the silhouette score SIL. For the ACM dataset, Fig. 4 (a) is $R = 3$ drawn from the node representations learned by UCHL, Fig. 4 (b) and Fig. 4 (c) obtained under the conditions of $R = 4$ and $R = 5$, respectively. It is observed that UCHL performs well on the ACM dataset and can identify clusters based on node representations. Changes in $R$ also do not cause large fluctuations in the quality of node representation, which can also be confirmed from the changes in SIL scores. In Figure 4, the SIL scores of UCHL in $R = 3$, $R = 4$ and $R = 5$ three cases are: 0.1185, 0.1114, and 0.1243, respectively . Under the dual constraints of mutual information and cluster center, UCHL can quickly aggregate similar node representations in the semantic vector space, and nodes in the same cluster are also close to the cluster center of the class.

# 4 Conclusion

This paper analyzes the influence of mutual information maximization and cluster center in the representation learning of heterogeneous graphs of scientific papers, and proposes an unsupervised cluster-level learning method of node representations of heterogeneous graphs of scientific papers (UCHL), which realizes the learning of heterogeneous graphs of scientific papers. Deep Semantic Representation Learning. UCHL uses the structure of meta-paths to model the semantics of associations in heterogeneous graphs, decomposes heterogeneous graphs into isomorphic graphs with specific semantics based on different meta-paths, and captures nodes with specific semantics based on graph convolutional neural networks local representation. Based on mutual information maximization and cluster clustering technology, UCHL uses an attention mechanism to aggregate node representations with different semantics. Good learning of heterogeneous graph node representations for scientific papers containing graph-level structural information. By validating the proposed UCHL on three scientific paper datasets, the experimental results show that the scientific paper heterogeneous graph node representation learned by UCHL achieves the best performance in the link prediction task of the scientific paper heterogeneous graph.

# References

[1] Wang X, Bo B, Shi C, et al. A survey on heterogeneous graph embedding: methods, techniques, applications and sources[J]. arXiv preprint arXiv:2011.14867, 2020.

[2] Li W, Xiao X, Liu J, et al. Leveraging graph to improve abstractive multi-document summarization[J]. arXiv preprint arXiv:2005.10043, 2020.

[3] Hussein R, Yang D, and Cudré-Mauroux P. Are meta-paths necessary? Revisiting heterogeneous graph embeddings[C]//Proceedings of the 27th ACM international conference on information and knowledge management, 2018: 437-446.

[4] Shi C, Han X, Song L, et al. Deep collaborative filtering with multi-aspect information in heterogeneous networks[J]. IEEE Transactions on Knowledge and Data Engineering, 2019, 33(4): 1413-1425.

[5] Li A, Du J, Kou F, et al. Scientific and Technological Information Oriented Semantics-adversarial and Media-adversarial Cross-media Retrieval[J]. arXiv preprint arXiv:2203.08615, 2022.

[6] Li W, Jia Y, Du J. Distributed extended Kalman filter with nonlinear consensus estimate[J]. Journal of the Franklin Institute, 2017, 354 (17): 7983-7995.

[7] Kong X, Philip S. Yu, Ying Ding, et al. Meta path-based collective classification in heterogeneous information networks[C]//Proceedings of the 21st ACM international conference on Information and knowledge management. 2012: 1567–1571.

[8] Perozzi B, Al-Rfou R, Skiena S. Deepwalk: Online learning of social representations[C]//Proceedings of the 20th ACM SIGKDD international conference on Knowledge discovery and data mining. 2014: 701-710.

[9] Grover A, Leskovec J. node2vec: Scalable feature learning for networks[C]//Proceedings of the 22nd ACM SIGKDD international conference on Knowledge discovery and data mining. 2016: 855-864.

[10] Narayanan A, Chandramohan M, Venkatesan R, et al. graph2vec: Learning distributed representations of graphs[J]. arXiv preprint arXiv:1707.05005, 2017.

[11] Li W, Jia Y, Du J. Recursive state estimation for complex networks with random coupling strength[J]. Neurocomputing, 2018, 219: 1-8.

[12] Wang D, Cui P, Zhu W. Structural deep network embedding[C]//Proceedings of the 22nd ACM SIGKDD international conference on Knowledge discovery and data mining. 2016: 1225-1234.

[13] Tang J, Qu M, Wang M, et al. Line: Large-scale information network embedding[C]//Proceedings of the 24th international conference on world wide web. 2015: 1067-1077.

[14] Chen H, Perozzi B, Hu Y, et al. Harp: Hierarchical representation learning for networks[C]//Proceedings of the AAAI Conference on Artificial Intelligence. 2018, 32(1).

[15] Meng D, Jia Y, Du J, et al. Data-driven control for relative degree systems via iterative learning[J]. IEEE Transactions on Neural Networks, 2011, 22(12): 2213-2225.

[16] Kipf T N, Welling M. Semi-supervised classification with graph convolutional networks[J]. arXiv preprint arXiv:1609.02907, 2016.

[17] Veličković P, Cucurull G, Casanova A, et al. Graph attention networks[J]. arXiv preprint arXiv:1710.10903, 2017.

[18] Hamilton W L, Ying R, Leskovec J. Inductive representation learning on large graphs[C]//Proceedings of the 31st International Conference on Neural Information Processing Systems. 2017: 1025-1035.

[19] Wang H, Leskovec J. Unifying graph convolutional neural networks and label propagation[J]. arXiv preprint arXiv:2002.06755, 2020.

[20] Zhang F, Bu T M. CN-Motifs Perceptive Graph Neural Networks[J]. IEEE Access, 2021, 9: 151285-151293.

[21] Fang Y, Deng W, Du J, et al. Identity-aware CycleGAN for face photo-sketch synthesis and recognition[J]. Pattern Recognition, 2020, 102: 107249.

[22] Luan S, Hua C, Lu Q, et al. Is Heterophily A Real Nightmare For Graph Neural Networks To Do Node Classification?[J]. arXiv preprint arXiv:2109.05641, 2021.

[23] Dong Y, Chawla N V, Swami A. metapath2vec: Scalable representation learning for heterogeneous networks[C]//Proceedings of the 23rd ACM SIGKDD international conference on knowledge discovery and data mining. 2017: 135-144.

[24] Shang J, Qu M, Liu J, et al. Meta-path guided embedding for similarity search in large-scale heterogeneous information networks[J]. arXiv preprint arXiv:1610.09769, 2016.

[25] Xue Z, Du J, Du D, et al. Deep low-rank subspace ensemble for multi-view clustering[J]. Information Sciences, 2019, 482: 210-227.

[26] Wang X, Ji H, Shi C, et al. Heterogeneous graph attention network[C]//The world wide web conference. 2019: 2022-2032.

[27] Fu X, Zhang J, Meng Z, et al. Magnn: Metapath aggregated graph neural network for heterogeneous graph embedding[C]//Proceedings of The Web Conference. 2020: 2331-2341.

[28] Zhang C, Song D, Huang C, et al. Heterogeneous graph neural network[C]//Proceedings of the 25th ACM SIGKDD International Conference on Knowledge Discovery & Data Mining. 2019: 793-803.

[29] Hu W, Gao J, Li B, et al. Anomaly detection using local kernel density estimation and context-based regression[J]. IEEE Transactions on Knowledge and Data Engineering, 2018, 32(2): 218-233.

[30] Hong H, Guo H, Lin Y, et al. An attention-based graph neural network for heterogeneous structural learning[C]//Proceedings of the AAAI Conference on Artificial Intelligence. 2020, 34(04): 4132-4139.

[31] Vaswani A, Shazeer N, Parmar N, et al. Attention is all you need[J]. Advances in neural information processing systems, 2017, 30.

[32] Hu Z, Dong Y, Wang K, et al. Heterogeneous graph transformer[C]//Proceedings of The Web Conference. 2020: 2704-2710.

[33] Velickovic P, Fedus W, Hamilton W L, et al. Deep Graph Infomax[J]. ICLR (Poster), 2019, 2(3): 4.

[34] Hjelm R D, Fedorov A, Lavoie-Marchildon S, et al. Learning deep representations by mutual information estimation and maximization[J]. arXiv preprint arXiv:1808.06670, 2018

[35] Li W, Jia Y, Du J. Variance-constrained state estimation for nonlinearly coupled complex networks[J]. IEEE Transactions on Cybernetics, 2017, 48(2): 818-824.

[36] Ren Y, Liu B, Huang C, et al. Heterogeneous deep graph infomax[J]. arXiv preprint arXiv:1911.08538, 2019.

[37] Mavromatis C, Karypis G. Graph InfoClust: Leveraging cluster-level node information for unsupervised graph representation learning[J]. arXiv preprint arXiv:2009.06946, 2020.

[38] Sun B, Du J, Gao T. Study on the improvement of K-nearest-neighbor algorithm[C]//2009 International Conference on Artificial Intelligence and Computational Intelligence, 2009: 390-393.

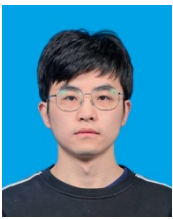

**Song Jie,** born in 1997, master. His main research interests include Data mining, information retrieval, machine learning.

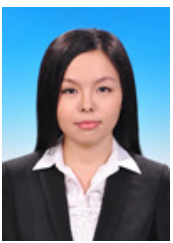

**Liang Meiyu** (corresponding author), was born in 1985. She is now an associate professor and master tutor at the School of Computer Science (National Pilot Software Engineering School), Beijing University of Posts and Telecommunications. Her main research directions are artificial intelligence, data mining, multimedia information processing, and computer vision.

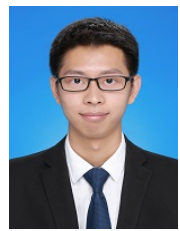

**Xue Zhe**, male, born in 1987, associate professor, master tutor. His main research interests are machine learning, artificial intelligence, data mining, and image processing. He has published more than 30 papers and 1 academic monograph. Presided over the National Natural Science Foundation of China Youth Fund project, participated in the national key research and development program projects, etc.

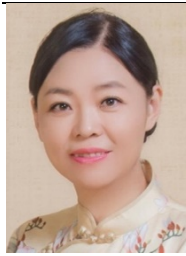

**Du Junping**, (1963 - ), female, professor, CCF Fellow ( NO.10411D ) . Main research areas are artificial intelligence, machine learning and pattern recognition

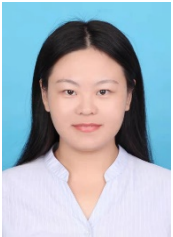

**Feifei Kou**, was born in 1989. She received her Ph.D. degree in School of Computer Science from Beijing University of Posts and Telecommunications, Beijing, China, in 2019. She ever did postdoctoral research in School of Computer Science from Beijing University of Posts and Telecommunications from 2019 to 2021. She is currently a lecturer in School of Computer Science (National Pilot Software Engineering School), Beijing University of Posts and Telecommunications, Beijing, China. Her research interests include semantic learning, and multimedia information processing.